\documentclass[a4paper]{article}

\usepackage{INTERSPEECH2021}
\usepackage[hidelinks]{hyperref}
\usepackage{enumitem}
\usepackage{tipa}
\usepackage{xcolor}

\newcommand\aspace{\hspace{.5em}}

\title{Phoneme Recognition through Fine Tuning of Phonetic Representations: a Case Study on Luhya Language Varieties}
\name{Kathleen Siminyu$^{\star}$ \aspace\aspace 
Xinjian Li$^{\dagger}$ \aspace\aspace 
Antonios Anastasopoulos $^{\ddagger}$  \\ 
\textit{David Mortensen$^{\dagger}$ \aspace\aspace 
Michael R. Marlo$^{\#}$ \aspace\aspace 
Graham Neubig$^{\dagger}$} }
\address{$^{\star}$Georgia Institute of Technology \qquad
      $^{\dagger}$Carnegie Mellon University \\
      $^{\ddagger}$George Mason University \qquad
      $^{\#}$University of Missouri}
\email{ksiminyu3@gatech.edu}

\begin{document}

\maketitle
\begin{abstract}
Models pre-trained on multiple languages have shown significant promise for improving speech recognition, particularly for low-resource languages. 
In this work, we focus on phoneme recognition using Allosaurus, a method for multilingual recognition based on phonetic annotation, which incorporates phonological knowledge through a language-dependent allophone layer that associates a universal narrow phone-set with the phonemes that appear in each language. 
To evaluate in a challenging real-world scenario, we curate phone recognition datasets for Bukusu and Saamia, two varieties of the Luhya language cluster of western Kenya and eastern Uganda. To our knowledge, these datasets are the first of their kind. We carry out similar experiments on the dataset of an endangered Tangkhulic language, East Tusom, a Tibeto-Burman language variety spoken mostly in India.
We explore both zero-shot and few-shot recognition by fine-tuning using datasets of varying sizes (10 to 1000 utterances). We find that fine-tuning of Allosaurus, even with just 100 utterances, leads to significant improvements in phone error rates.\footnote{Our fine-tuning code and data will be released upon publication.}
\end{abstract}

\section{Introduction}

While speech recognition has made great strides, ASR over languages with very little transcribed text is still a daunting task.
One of the most promising directions in low-resource speech processing involves pre-training models, both to improve low-resource ASR itself and to use ASR as an auxiliary task to improve results in other low-resource task settings.
For example, Wiesner et al.~\cite{wiesner2018pretraining} explore pre-training by back-translation for end-to-end ASR, while Stoian et al.~\cite{stoian2020analyzing}, Bansal et al.~\cite{bansal2018pre} use ASR as an auxiliary task to improve results for low-resource speech-to-text translation. Recently, Baevski et al.~\cite{baevski2020wav2vec} introduced wav2vec 2.0, which can be used to achieve good ASR performance by fine-tuning on as little as 10 minutes of transcribed audio.
Further, pre-training has been combined with multi-lingual learning; Wang et al.~\cite{wang2020improving} improve cross-lingual transfer learning for ASR with speech translation, while Hsu et al.~\cite{hsu2020meta} formulate ASR for different languages as different tasks and use a model agnostic meta-learning algorithm (MAML) to learn the various initialization parameters.

Nonetheless, effective multilingual learning of acoustic and language models is difficult, given the underlying differences between languages in pronunciation and lexicon respectively.
To help alleviate issues caused by pronunciation differences, Li et al.~\cite{li2020universal} have recently proposed a model (Allosaurus; details in \S\ref{sec:methodology}), that calculates a language-universal phone distribution using a standard ASR encoder, then converts it to a language-specific phoneme distribution. \emph{Phonemes} are sounds that support lexical contrasts in a \emph{particular} language;  \emph{phones} are the sounds that are physically spoken (which are largely language independent); and \emph{allophones} are the set of phones that correspond to a particular phoneme.
This method showed promise both for recognition on the datasets on which the model was pre-trained, and also for \emph{zero-shot} adaptation to new languages, where it was tested on languages for which no training data was available.

However, in realistic low-resource settings, it is common to have a \emph{small amount} of training data in the language to which we would like to adapt.
Our work examines the question: ``given a pre-trained acoustic model and phone recognizer based on universal phone representations, how quickly can it be adapted to perform reasonable phone recognition in a new language?''
To examine this question, we perform both zero-shot recognition and fine-tuning using datasets of sizes varying between 10 and 1000 utterances.
As an important second contribution, we curate phone recognition datasets for Bukusu and Saamia, two varieties of Luhya, a cluster of Bantu languages largely spoken in Kenya and Uganda; to our knowledge these are the first datasets of their type.
Using these datasets we perform fine-tuning experiments and find both respectable zero-shot results, and rapid improvements with very small amounts of adaptation, demonstrating the utility of fine-tuning of universal phonetic representations as a method for building very low-resource ASR models. We also carry out similar experiments on the dataset of an endangered Tangkhulic language, East Tusom, a Tibeto-Burman language variety spoken mostly in India.

\noindent Briefly summarized, with this work we:
\begin{itemize}[noitemsep,nolistsep,leftmargin=*]
    \item provide a phoneme recognition benchmark on Bukusu and Saamia, two under-resourced Luhya language varieties, as well as East Tusom, an endangered Tibeto-Burman language variety,
    \item show that fine-tuning of Allosaurus can lead to phone error rate (PER) reductions of more than 40\%, and 
    \item show that few-shot fine-tuning using only 100 transcribed utterances can lead to up to 59\% PER reductions 
\end{itemize}

\section{Methodology}
\label{sec:methodology}

\begin{figure}[t]
  \centering
  \includegraphics{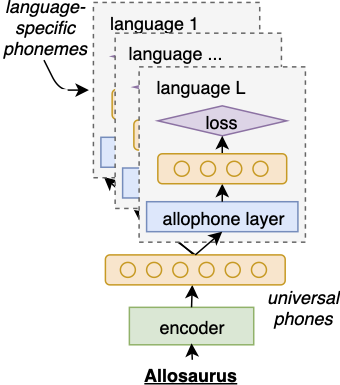}
  \caption{
  Allosaurus (a multilingual pre-trained phoneme recognition model) predicts over a shared phone inventory, then maps into language-specific phonemes with an allophone layer.}
  \vspace{-5mm}
\label{arch}
\end{figure}

\subsection{Allosaurus Layer}

\emph{Allosaurus},  (Figure \ref{arch}; \cite{li2020universal}), comprises a language independent encoder and phone predictor, and a language dependent allophone layer and a loss function associated with each language. This combination makes Allosaurus appropriate for multilingual phoneme recognition, because it allows it to handle phonemes while modeling the underlying phones, unlike other traditional multilingual models.
The encoder first produces the distribution $h \in \mathbb{R}^{|P_{\text{uni}}|}$ over the universal phone inventory $P_{\text{uni}}$, then the allophone layer transforms $h$ into phoneme distribution $g^i \in \mathbb{R}^{|Q_{i}|}$ of each language. 
The allophone layer uses a trainable allophone matrix $W^i \in \mathbb{R}^{|Q_i| \times |P_{\text{uni}}|}$ to describe allophones in a way similar to the signature matrix $S^i = \{0,1\}^{|Q_i| \times |P_{\text{uni}}|}$ which describes the association of phones and phonemes in each language $L_i$. The allophone matrix $W^i$ is first initialized with  $S^i$, and is allowed to be optimized during the training process. An L2 penalty is added to penalize divergence from the original signature matrix $S^i$. The allophone layer computes its logit distribution $g^i$ by finding the most likely allophone realization in $P_{\text{uni}}$ with maxpooling. 

\begin{equation}
g^i_{j}=\max(\{ w^i_{j,k} \cdot h_{k} ; 1 \leq k \leq |P_{\text{uni}}| \}),
\end{equation}
where $g^i_{j} \in \mathbb{R}$ is the logit of $j$-th phoneme in $g^i$ for language $L_i$, $w^i_{j,k} \in \mathbb{R}$ is the $(j,k)$ cell of the allophone matrix $W^i$, $h_{k} \in \mathbb{R}$ is the logit of $k$-th phone in $h$. Intuitively, if the $j$-th phoneme has the $k$-th phone as an allophone, $w^i_{j,k}$ would be near 1, otherwise $w^i_{j,k}$ would be near 0. Therefore, the phoneme logit of $g^i_j$ is decided by the largest allophone logit $h_k$. The phoneme distribution $g^i$ is further fed into the loss function. This method for phoneme prediction can be used with any underlying multilingual ASR system. Here the parameters are optimized by minimizing CTC loss, Graves et al. \cite{graves2006connectionist}, for all training languages, with the addition of regularization of the allophone layer controlled by hyperparameter $\alpha$.

\begin{equation}
\mathcal{L}=\sum_{1 \leq i \leq |L|} (\mathcal{L}^i_{ctc} + \alpha \lVert W^i-S^i \rVert_2^2 ).
\vspace{-2mm}
\end{equation}

\subsection{Universal Phone Recognition}

Not only does the allophone layer abstract away from the language-specific phonemes, which contributes to the improvement in the multilingual acoustic modeling, the model also gives us the capability to predict universal phones themselves.
By applying a greedy decoding strategy over the phone distribution $h$, we can obtain a phone sequence in which all phones $P_{\text{uni}}$ in the training languages are candidates. Combined with large training language sets, the universal inventory covers most common narrow phones appearing in most languages. 

Furthermore, this protocol can take into account phone inventories that have already been created for many languages in the world by linguists.
For example, PHOIBLE \cite{moran2019phoible} is a database of phone inventories for more than 2000 languages and dialects, allowing our model to be applied to these languages with some degree of accuracy.
If the phone inventory for language $L_i$ is $P_i$, we can restrict the decoder to only produce phones in $P_i \cap P_{\text{uni}}$ by filtering out other phones.
When the universal inventory $P_{\text{uni}}$ covers most frequent phones in the world, we could expect that $P_i \approx P_i \cap P_{\text{uni}}$.

\subsection{Fine-tuning of Universal Phonetic Representations}

In Li et al.~\cite{li2020universal}, the universal phonetic representations were tested in a \emph{zero-shot} setting, where the model was used as-is on languages not occurring in the training data.
However, in many cases a small amount of labeled data \emph{does} exist in the target language we wish to recognize.
In this paper, we further propose and demonstrate results for \emph{fine-tuning of universal phone representations} as an efficient and expedient way to utilize this data.
Depending on the type of transcriptions that are available for the new language (phonetic or phonemic), we can fine-tune either using only the shared universal phone output layer, or create a new allophone layer for the new language and use a language-specific loss.
In this work, we fine-tune the encoder which produces a distribution over the universal phone inventory. 

We fine-tune with a small learning rate, 0.01, to avoid catastrophic forgetting~\cite{french1999catastrophic} for a maximum of 250 epochs, and choose the best performing model based on a small validation set. We use simple stochastic gradient descent as the optimization algorithm. 

\section{Phoneme Recognition Benchmark for Luhya Language Varieties}
\label{sec:benchmark}

In this section we describe the background and the data sources for our Luhya phoneme recognition benchmarks and the data source for the East Tusom benchmark.
\vspace{-.3cm}

\subsection{Background}
Bukusu and Saamia are members of the Luhya cluster of $\sim$25 Bantu languages. Though Luhya languages are spoken in both western Kenya and eastern Uganda, ``Luhya" typically refers to the $\sim$19 Kenyan communities that were politically united in the first half of the 20th century.\footnote{See MacArthur \cite{macarthur_cartography_2016} for a recent statement about modern Luhya history.} Luhya is used as a label of both ethnicity and language for this group of culturally, politically, and linguistically heterogeneous communities. 

Together, Kenyan Luhya communities number around 6.8 million people and are the second-largest ethnic group in Kenya.\footnote{Details here: \href{https://www.knbs.or.ke/?wpdmpro=2019-kenya-population-and-housing-census-volume-iv-distribution-of-population-by-socio-economic-characteristics&wpdmdl=5730&ind=7HRl6KateNzKXCJaxxaHSh1qe6C1M6VHznmVmKGBKgO5qIMXjby1XHM2u_swXdiR}{2019 Kenya Census}. Retrieved March 2020.} 
Bukusu is the largest Kenyan Luhya community with more than 1 million members; there are more than 85,000 members of the Saamia community in Kenya, and another 279,000 in Uganda.\footnote{Details here: \href{https://www.ubos.org/wp-content/uploads/publications/03_20182002_CensusPopnCompostionAnalyticalReport_(1).pdf}{2002 Uganda Census}. Retrieved October 2020.}

Marlo et al.\ \cite{Marlo-et-al:2020} provide a recent classification of Luhya languages, showing that despite their diversity, the Luhya language varieties are indeed more closely related to one another than they are to their nearest Bantu neighbors to the west and south. All modern linguistic studies such as Mutonyi \cite{mutonyi_aspects_2000} and Botne et al.\ \cite{botne_grammatical_2006} on Saamia focus on the languages of individual sub-communities, who are each recognized to have their own distinct speech form. Whether each language variety should be considered as a ``dialect" or a ``language" is a matter of debate, but see Angogo \cite{angogokanyoro1983} for a study of mutual intelligibility among Kenyan Luhya varieties.

Early missionary activity introduced competing orthographies for Luhya languages. Despite some standardization efforts, there are not universally agreed upon orthographic conventions for writing all Luhya varieties, which are linguistically diverse and have different sound systems.
\vspace{-.3cm}

\subsection{Access to Data}
\label{ssec:dataaccess}
An inescapable aspect of working in low-resource languages is identifying suitable data. 

The CMU Wilderness Multilingual Speech Dataset covers over 700 languages providing audio, aligned text and word pronunciations. Saamia, one of the Luhya varieties, is available in the CMU Wilderness dataset. On average each language has around 20 hours of sentence-length transcriptions. Data is mined from reading of the New Testament from Bible.is.\footnote{\href{http://www.bible.is/}{Bible.is website}}

The co-author Michael Marlo, who has ongoing documentation projects on several Luhya language varieties, also provided access to recordings from his unpublished dictionary of Bukusu. Most recordings include two or more pronunciations of the same word. Similarly, the East Tusom dataset, composed of transcribed audio from a comparative wordlist was made available by the co-author David Mortensen through his documentation work of the language.

\begin{table*}[t]
\centering
\caption{Fine-tuning of Allosaurus (even with just 100 utterances) leads to significant improvements in PER. The PER improvements ($\Delta$) reported are a percentage relative to the language-constrained baseline.}
\vspace{-.3cm}
\label{table}
\begin{tabular}{l|cc|cc|cc}
\toprule
  & \multicolumn{2}{c|}{Bukusu} & \multicolumn{2}{c|}{Saamia} & \multicolumn{2}{c}{East Tusom} \\
  & PER & $\Delta$ & PER & $\Delta$ & PER & $\Delta$ \\
 \midrule
 \multicolumn{1}{l|}{Allosaurus} & 72.8 & & 63.7 & & 67.5\\ 
 \ \ + constraint & 52.5 & & 37.4 & & 56.7\\ 
 \midrule
 \ \ + fine-tuning (100) & 41.2 & 21.5\% & 15.5 & 58.5\% & 44.8 & 20.9\%\\
 \ \ + fine-tuning (1000) & 17.3 & 67.0\%& 11.7 & 65.7\% & 34.6 & 38.9\%\\
 \ \ + fine-tuning (all) & 5.2 & 90.1\%& 9.2& 75.4\% & 33.1 & 41.6\%\\ 
 \bottomrule
\end{tabular}
\end{table*}

\subsection{Data Preparation}
\label{ssec:dataprep}

\noindent\textbf{Saamia Data \ }
The CMU Wilderness GitHub repository\footnote{\href{https://github.com/festvox/datasets-CMU_Wilderness}{CMU Wilderness GitHub repository}} contains code to download data directly from the Bible.is website as they do not have permission to redistribute this data. It contains 18.2 hours of Saamia data. Alignments, short waveforms plus transcripts, can then be reconstructed for each language from a packed version contained in the GitHub repository. 

\noindent\textbf{Bukusu Data \ }
The dictionary recordings contain a word uttered several times in various forms, e.g.~noun recordings include variations of the headword, singular and plural forms of the word while verbs include variations of the headword, infinitive and stem. The recordings also include pronouns, adjectives, adverbs, numerals and conjunctions. The dictionary contains 3.7 hours of data. 
The preparation process for these recordings included using SoX\footnote{\href{http://sox.sourceforge.net/}{SoX-Sound eXchange}} to change the sampling rate and in some cases the sampling encoding. We used the WebRTC\footnote{\href{https://github.com/wiseman/py-webrtcvad}{python interface to the WebRTC Voice Activity Detector}} voice activity detection tool to split the audios on silences, and manually inspected the result to ensure all segments and transcriptions matched up properly.

\noindent\textbf{East Tusom Data \ }
This recently published dataset, Tusom2021\footnote{\href{https://github.com/dmort27/tusom2021/tree/main/data}{East Tusom Github repo}}, was prepared explicitly for the task of phone recognition and therefore needed no pre-processing. It contains 55.3 minutes of audio data.

\noindent\textbf{Orthography-to-Phone Mapping \ }
\label{sssec:phonephonememapping}
While the Bukusu and East Tusom transcriptions were already in the International Phonetic Alphabet (IPA), it was necessary to develop an orthography to phone (IPA) mapping for Saamia to enable us obtain phonetic transcriptions of our data using Epitran \cite{Mortensen-et-al:2018}. These mappings were developed in collaboration with linguists and will be released as additions to the Epitran IPA transcriber~\cite{Mortensen-et-al:2018}.

\section{Experiments}
\label{sec:experiments}

The experimental splits used for the datasets were as follows:
\begin{itemize}[noitemsep,nolistsep,leftmargin=*]
    \item Bukusu: 6442 (train), 1001 (dev), 2458 (test)
    \item Saamia: 7254 (train), 1000 (dev), 1500 (test)
    \item East Tusom: 1600 (train), 400 (dev), 392 (test)
\end{itemize}

\subsection{Experimental Setting}
Given a pre-trained acoustic model and phone recognizer based on universal phone representations, we are interested in understanding how much data leads to improvements in model performance via fine-tuning. We thus create datasets of different sizes; 10, 25, 50, 100, 250, 500 and 1000. This selection of dataset sizes is intended to be an approximately doubling progression. These datasets are transcribed with phones, we therefore evaluate using phone error rate during fine-tuning of the languages and use the same test set across different dataset sizes for each language.
All fine-tuning is done on one model therefore using the same encoder, a 6-layer stacked bidirectional LSTM with hidden size of 1024 in each layer. Fine-tuning in each instance is carried out for 250 epochs, beyond which we noted no significant variation. 

\subsection{Experimental Results}

Table \ref{table} shows top-line PER results of inferences in various conditions; the results of zero-shot inference using the pre-trained Allosaurus model, the results when the allophone layer is constrained to phones found in the respective languages, the best results after fine-tuning using a 100-sized dataset, the best results after fine-tuning using a 1000-sized dataset and the results after fine-tuning using the entire training set of each language dataset.
We deduce that fine-tuning, even with just 100 utterances, leads to significant improvements in PER.

\begin{figure}[t!]
  \centering
  \centerline{\includegraphics[width=1\columnwidth]{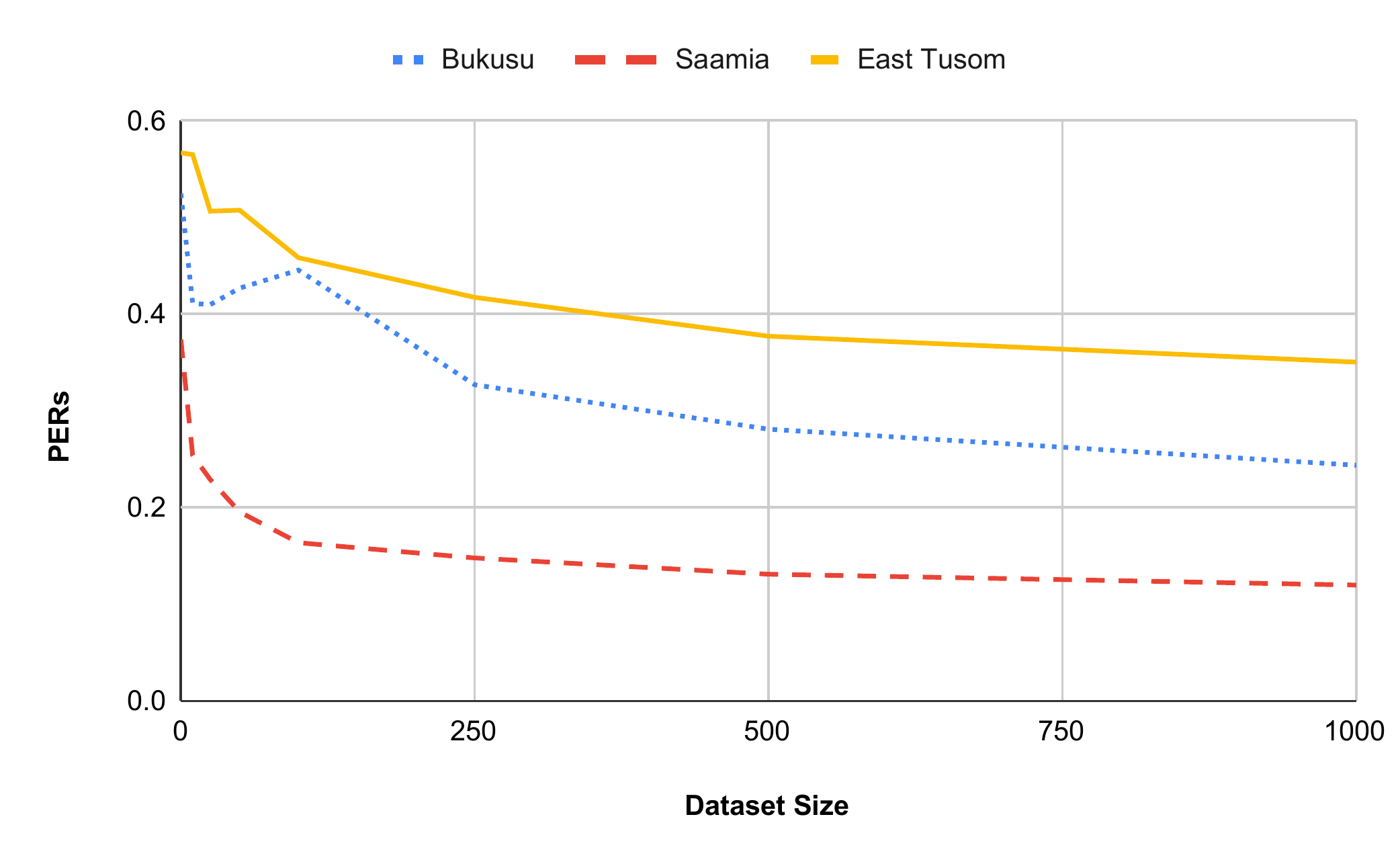}}
  \vspace{-0.3cm}
  \caption{Dataset sizes against PERs curves at training epoch 40 for Bukusu, Saamia and East Tusom.}
  \vspace{-.3cm}
  \label{fig2}
  \vspace{-.3cm}
\end{figure}

Figure \ref{fig2} shows the variation of PERs across various dataset sizes at training epoch 40 for all 3 languages. We note that as expected, the PER mostly decreases as dataset size increases, particularly for Saamia. 

\begin{figure}[t!]
\begin{minipage}[b]{1.0\linewidth}
  \centering
  \centerline{\includegraphics[width=8cm]{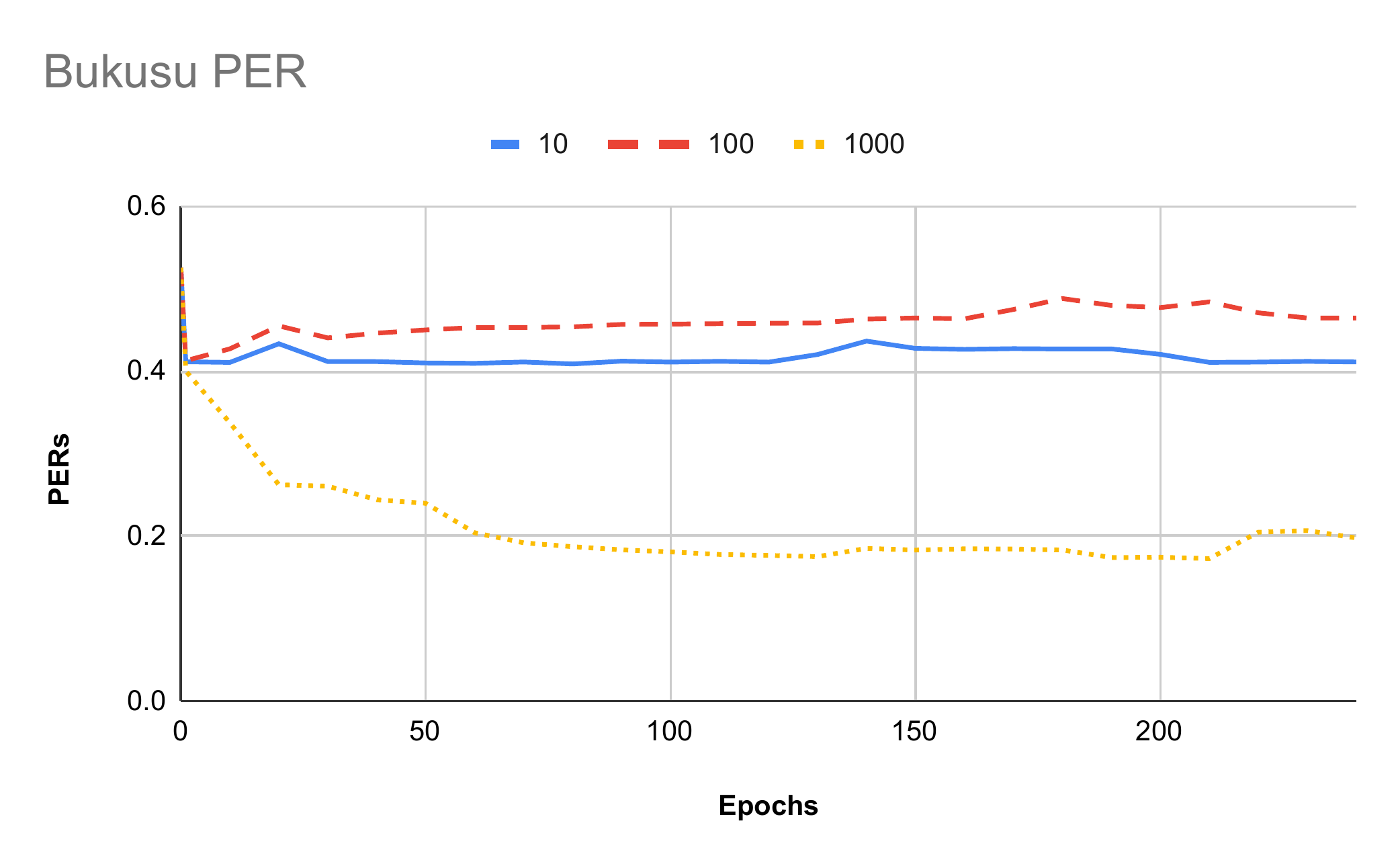}}
  \vspace{-0.3cm}
  \centerline{\includegraphics[width=8cm]{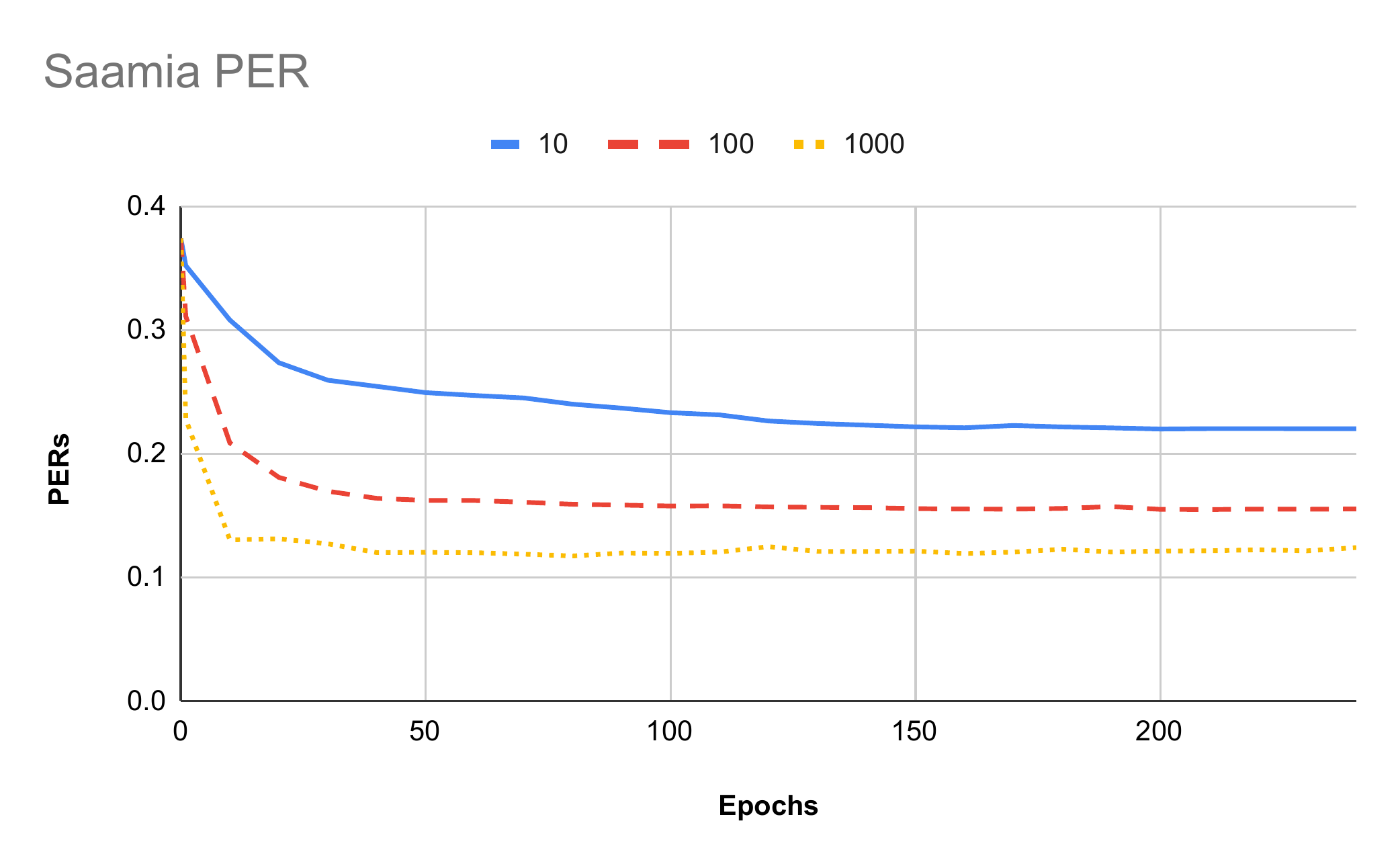}}
  \vspace{-0.3cm}
  \centerline{\includegraphics[width=8cm]{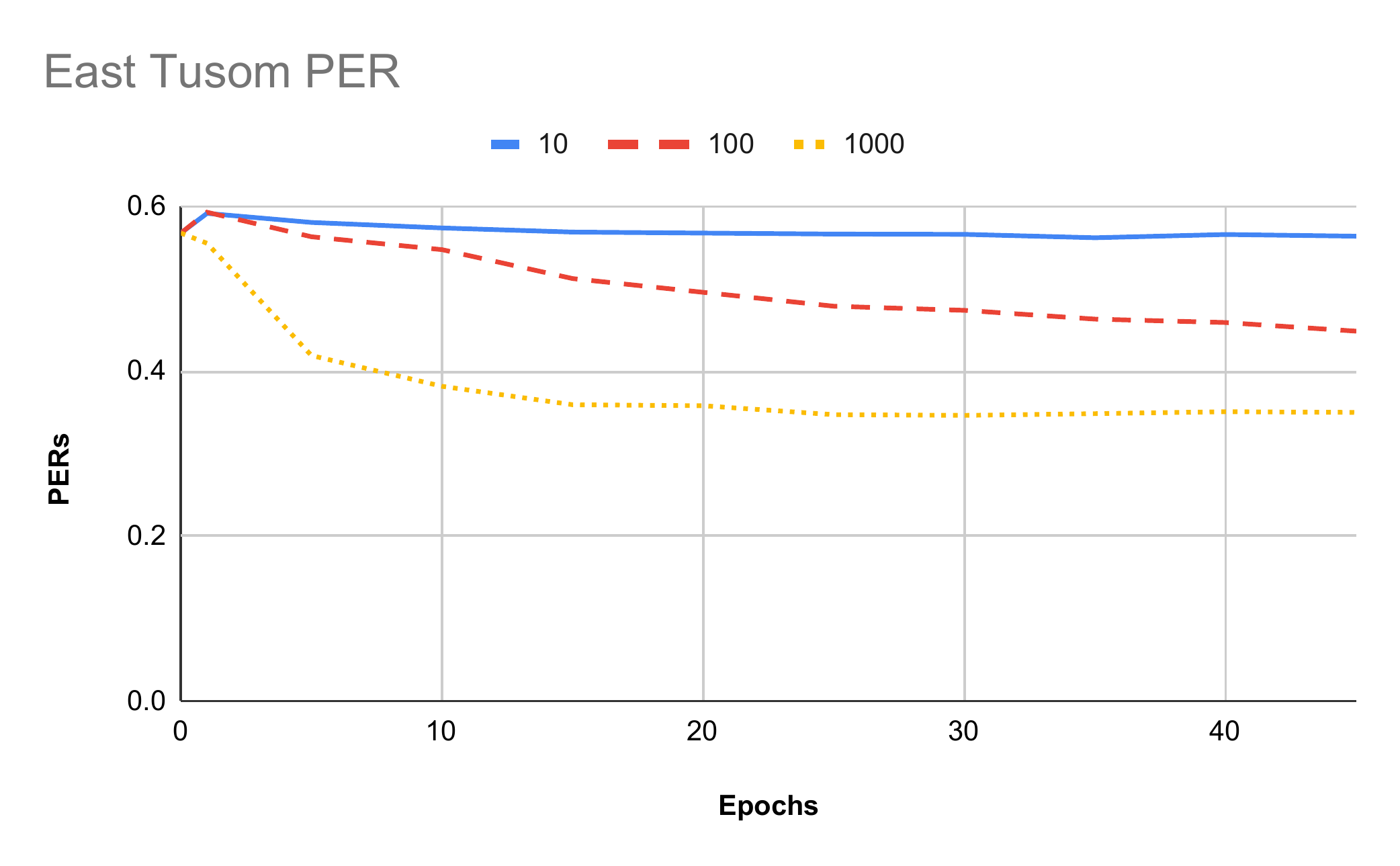}}
\end{minipage}
\vspace{-0.7cm}
  \caption{Bukusu, Saamia and East Tusom accuracy over training epochs for dataset sizes 10, 100 and 1000. Due to early stopping in the case of East Tusom, we plot over 45 epochs in that instance and over 250 epochs for the others.}
  \label{figures3}
   \vspace{-0.8cm}
\end{figure}

Figure \ref{figures3} shows the accuracy over training epochs for dataset sizes 10, 100 and 1000. Across all varieties (and when an adequate number of training instances are available) fine-tuning quickly improves the PER after a few epochs. We note that smaller fine-tuning datasets, of just 10 or 100 instances in Bukusu and 10 instances in East Tusom, only require a handful of fine-tuning epochs to reach their best performance.
In the case of Bukusu, Figure \ref{figures3}(top) makes apparent that the 10 to 100 dataset size PERs are somewhat consistent, even appearing to worsen as fine-tuning progresses. This is likely due to the nature of the data, where each training instance is a single dictionary recording and each audio contains one word. This means the dataset size is equivalent to the number of words or utterances in the training set, so dataset sizes 10 to 100 contain very little data. East Tusom exhibits behaviour similar to Bukusu as shown in Figure \ref{figures3}(bottom) likely due to the fact that the dataset is also composed of individual utterances. We see that PERs for the 10 and 100-sized datasets in some of the initial training epochs become slightly worse than the zero-shot results with the language-constrained baseline before improving. In the case of the 10-sized dataset, the improvements over zero shot results with the language-constrained baseline are negligible, however when the datasets get bigger, we start to see better improvements with few training epochs before getting to the point of early stopping, 45 epochs. 


A manual inspection of the differences of the test set outputs of the baseline and the fine-tuned models 
reveal interesting patterns. The Bukusu and Saamia baseline models under-perform on long vowels, outputting short ones; the Bukusu fine-tuned model corrects this mistake in over 400 cases (e.g. the \textipa{i}$\rightarrow$\textipa{i:} mistake is corrected~119 times) and the Saamia one in over 600 cases(e.g. the \textipa{a}$\rightarrow$\textipa{a:} mistake is corrected~314 times). The Saamia model also under-performs on consonant blends(e.g. the \textipa{d}$\rightarrow$\textipa{nd} mistake is corrected~294 times and \textipa{n}$\rightarrow$\textipa{nd} mistake is corrected~233 times).  The Bukusu baseline model also seems to confuse the liquids (\textipa{l}, \textipa{R} and \textipa{r}) an issue that fine-tuning largely addresses. 
On the other hand, the fine-tuned model over-recognizes long vowels, and it seems slightly more likely to drop or insert phonemes spuriously.

In the case of East Tusom, the baseline model appears to struggle with differentiating the vowels with many of the corrections made by the fine-tuned model being replacing one vowel with another. (eg. \textipa{o}$\rightarrow$\textipa{u}, \textipa{a}$\rightarrow$\textipa{o}, \textipa{e}$\rightarrow$\textipa{i}, \textipa{a}$\rightarrow$\textipa{\~o})

\vspace{-0.3cm}
\section{Implications and Future Work}
We show that fine-tuning of the Allosaurus model is a promising approach to phone recognition for unseen low-resource languages.
The results indicate improvements over the language-constrained baseline with as little as 10 training instances (of course, using more fine-tuning data leads to even further improvements). 

Future work could include tuning of the parameters of the Allosaurus model itself, or more sophisticated training techniques such as layer-by-layer adaptation of the encoder or allophone layers~\cite{howard2018universal} and/or use of meta-learning~\cite{hsu2020meta}.
In addition, as some language pairs are more closely related than others, it might be interesting to experiment with further fine-tuning on multiple languages from a language group and with mixed fine-tuning~\cite{neubig-hu-2018-rapid} where we sample some data from the original data and some from the target language.
Finally, as many languages, such as the Luhya languages we covered here, are tonal it would be interesting to include tones in the pretraining of Allosaurus and perform tonal recognition.  
Finally, while we focus on phone recognition here, it would be of great utility to incorporate similar techniques into a full-fledged speech recognition systems.

\bibliographystyle{IEEEtran}

\bibliography{mybib}


\end{document}